# Novel Feature-Based Clustering of Micro-Panel Data (CluMP)

Lukáš Sobíšek[1], Mária Stachová[2], Jan Fojtík[1]

**Corresponding Author:**

Lukáš Sobíšek,

email: lukas.sobisek@yahoo.com,

[1]Faculty of Informatics and Statistics, University of Economics,

nám. W. Churchilla 1938/4, 130 67 Praha 3, Prague, Czech Republic

**Contacts:**

Mária Stachová,

email: maria.stachova@umb.sk,

[2]Faculty of Economics, Matej Bel University,

Tajovskeho 10, 975 90, Banska Bystrica, Slovakia

Jan Fojtík,

email: xfojj00@vse.cz,

[1]Faculty of Informatics and Statistics, University of Economics,

nám. W. Churchilla 1938/4, 130 67 Praha 3, Prague, Czech Republic

**acknowledgment:**

Lukáš Sobíšek and Jan Fojtík has been supported by the project of the University of Economics, Prague - Internal Grant Agency, project No. 44/2017 "Clustering and regression analysis of micro panel data".

Mária Stachová has been supported by the PROJECT VEGA NO. 1/0093/17 Identification of risk factors and their impact on products of the insurance and saving schemes.

# Abstract

Micro-panel data are collected and analysed in many research and industry areas. Cluster analysis of micro-panel data is an unsupervised learning exploratory method identifying subgroup clusters in a data set which include homogeneous objects in terms of the development dynamics of monitored variables. The supply of clustering methods tailored to micro-panel data is limited. The present paper focuses on a feature-based clustering method, introducing a novel two-step characteristic-based approach designed for this type of data. The proposed CluMP method aims to identify clusters that are at least as internally homogeneous and externally heterogeneous as those obtained by alternative methods already implemented in the statistical system R. We compare the clustering performance of the devised algorithm with two extant methods using simulated micro-panel data sets. Our approach has yielded similar or better outcomes than the other methods, the advantage of the proposed algorithm being time efficiency which makes it applicable for large data sets.

## 1. Introduction

Panel longitudinal studies are conducted primarily for the purpose of analysing value changes in monitored variables over time. According to the sample type, the panel data set can be broken down into micro- or macro-panels. The micro-panel is a set containing lots of statistical objects (usually hundreds or thousands of them) that are periodically observed over time, the numbers of repeated measurements being significantly smaller. Let us call the sequence of $T$ repeated measurements of a variable for the $i$-th out of $N$ micro-panel objects a trajectory; it can be viewed as a curve. Micro-panel data are used in various disciplines. Examples include a cross-national panel database of micro-data on health, socioeconomic status and social and family networks (SHARE) or the Czech registry of Patients with Multiple Sclerosis (ReMuS Registry). In the area of finance, we have worked with micro-panel data for the prediction of financial distress of companies (Sobíšek et al., 2017; Stachová et al., 2017). More often, panel data are employed in non-economic areas such as medicine, education, psychology, political science, ecology or zoology. In the field of medicine, we have used micro-panel data (patients' cohort), for example, to identify biomakers of disability development in patients with multiple sclerosis (Uher et al., 2017) or to predict an individual treatment response (Kalinčík et al., 2017). A macro-panel contains fewer objects (in units or tens) that are compared to each other over more time observations (at least in the order of dozens). The sequence of $T$ repeated measurements for each $i$-th macro-panel object is a time series.

Univariate or multivariate cluster analysis of micro-panel trajectories is a common unsupervised learning exploratory method identifying subgroup clusters in a data set which include homogeneous objects in terms of the development dynamics of monitored variables (trajectories). There are various reasons for clustering of objects according to the dynamic development of the analyzed one or more variables. In exploratory analysis, the objective may be to identify outlier trajectories or clusters of outliers that can be removed e.g. from regression analysis or examined separately. Clustering algorithms are also useful for finding representative curves corresponding to different modes of variation (Tarpey and Kinateder, 2003). In this sense, trajectoral clustering can also be a primary goal of analysis, for example, in the areas of public healthcare (Prochaska et al., 1991), epidemiology (Koestler et al., 2014) and economics (Bartošová and Longford, 2014).

There are relatively many methods for clustering macro-panel data (time series). An overview of three different approaches to cluster time series is provided by Liao (2005) who gives examples of each of them:

(1) *Model-based approaches* emphasize that each time series is generated by some kind of model or a mixture of underlying probability distributions. Modifications of Gaussian processes for the clustering of time series are proposed by Koestler et al. (2014), De la Cruz-Mesía et al. (2008), Komárek (2009), Bouveyron and Jacques (2011), Gattone and Rocci (2012) and Yamamoto and Hwang (2017).

(2) *Raw-data-based approaches* work with raw data. These methods modify clustering algorithms originally developed for static cross-sectional spatial data. Such an adjustment is proposed, e.g. by Lombardo and Falcone (2011).

(3) *Feature/characteristic-based approaches* are data mining algorithms using clustering of extracted features of time series, characteristics derived from the original values being clustered. Transformation suited for time series with multiple time iterations (macro-panel) is summed up by Liao (2005) or Wang et al. (2006).

On the other hand, the offer of micro-panel data clustering methods in the R statistical system is significantly reduced (visit www.r-project.org). Moreover, we found out that the methods implemented in mixAK and KML packages (Komárek, 2009; Genolini and Falissard, 2011, respectively) did not better recognize existing patterns of the development of repeated measurements in time (Stachová et al., 2017; Stachová and Sobíšek, 2016). Also, with a large number of clustered objects (in the order of thousands), the calculation speed of these algorithms was rather slow (in minutes). For big data, these functions seem to be uncorverted. For the above reasons, the aim of this study is to describe our own, time-efficient micro-panel data clustering procedure.

The present paper focuses on the feature-based clustering method, introducing a novel two-step characteristic-based approach designed for micro-panel data clustering, called CluMP (Clustering of Micro Panel).

In the first step, the panel data are transformed into static data with lower dimension using a set of the proposed dynamic characteristics, representing different features of the time course of the observed variables. In the second step, the elements are clustered by clustering techniques designed for static data.

Using a simulation study, we compare the clustering capability of the CluMP algorithm with that of two alternative algorithms implemented in the R statistical system. The algorithm represents the feature-based technique. From the model-based method, we have chosen the mixAK algorithm (Komárek, 2009), the raw-data-based approach representing KML functions (Genolini and Falissard, 2011).

The paper has the following structure. In the next section, we define the feature-based approach. Section 3 is devoted to the simulation study overview, namely technical specification and data and evaluation criteria description. The simulation study results for balanced data and those for unbalanced simulated data are presented in Section 4 and 5, respectively. The conclusions of the research are summarized at the end of the paper. In the appendix section, an example illustrating the calculation of trajectory feature data (the first clustering stage) are shown. The R code is available upon request to the corresponding author.

## 2. Feature-based clustering approach (CluMP)

The present study describes a two-step algorithm for clustering micro-panel trajectories (CluMP). In his dissertation thesis, Sobíšek (2017) proposes six combinations of characteristics that represent different properties of the short-time development of monitored variable values ($i$-th trajectory). For each combination, seven clustering algorithms (i.e. six agglomerative algorithms and the $C$-means partitioning one) were applied. Having used the simulation study, the best combination of transforming characteristics and the most appropriate clustering algorithm applied were selected. They are presented in Sections 2.1 and 2.2, respectively.

### 2.1 Data transformation features

In the present algorithm, we identify our own set of seven features describing the dynamics of individual trajectories, providing a specific nonparametric description of trend and variation of repeated measurements. The suggested characteristics include:

*Average triangular difference between the two consecutive measurement values*

For the $i$-th object, the average value is denoted $\overline{diff}_i$. It is calculated as the mean of all triangular differences between the two consecutive time points $diff_{it}$ as follows

$$\overline{diff}_i = \frac{\sum_{t=2}^{T} diff_{it}}{T-1} = \frac{\sum_{t=2}^{T} 1/2 \left( \frac{y_{it} - y_{i(t-1)}}{D_{it} - D_{i(t-1)}} \right)}{T-1}, \tag{1}$$

where $y_{it}$ is the value of the monitored variable for the $i$-th object in time $t$, $D_{it}$ denotes the time difference of the time point $t$ from the beginning ($t = 1$) for the $i$-th object (e.g., in years) and $T$ is the number of observations of the $i$-th object.

*Selective standard deviation of triangular differences between the two consecutive measurements*

For the *i*-th object, the variability characteristic, denoted sd(*diff$_i$*), is calculated as

$$\text{sd}(\textit{diff}_i) = \frac{\sum_{t=2}^{T}\left(\textit{diff}_{it} - \overline{\textit{diff}_i}\right)^2}{T-2}. \tag{2}$$

*Average absolute triangular defference between the two consecutive measurements*

For the *i*-th object, this absolute value, denoted $\overline{|\textit{diff}_i|}$, is calculated as

$$\overline{|\textit{diff}_i|} = \frac{\sum_{t=2}^{T}|\textit{diff}_{it}|}{T-1}. \tag{3}$$

Compared to (1), absolute values of triangular differences between $y_{it}$ a $y_{i(t-1)}$ are averaged.

*Selective standard deviation of absolute triangular differences between the two consecutive measurements*

For the *i*-th object, the standard deviation, denoted $\text{sd}(|\textit{diff}_i|)$, is calculated as

$$\text{sd}(|\textit{diff}_i|) = \frac{\sum_{t=2}^{T}\left(|\textit{diff}_{it}| - \overline{|\textit{diff}_i|}\right)^2}{T-2}. \tag{4}$$

*Average growth coefficient $\overline{k_i}$*

The mean growth rate indicates the average relative change of the monitored variable *Y* and is calculated for the *i*-th object as

$$\overline{k_i} = \sqrt[n_i-1]{\frac{y_{iT}}{y_{i1}}} - 1 = \sqrt[n_i-1]{k_{i2}k_{i3}...k_{it}...k_{iT}} - 1, \qquad (5)$$

where $k_{it} = y_{it}/y_{i(t-1)}$ for $t = 2, ..., T$ denotes the coefficient of growth between periods $t$ and $t - 1$ of the variable $Y$.

*The ratio of positive to negative changes (%pos)*

$$\% pos = \frac{\text{number\_}k_{it} \geq 1}{\text{number\_}k_{it} \leq 1}, \quad t = 2,...,T, \qquad (6)$$

if the number of negative changes (denominator) is zero, the value in the denominator is replaced with a value of 0.1.

*The value of maximum angle between the line connecting peripheral measurements and the one between the inner point and the first measurement (in radians)*

The maximum angle is given in radians as an angle between the line connecting peripheral measurements [$t = 1$, $y_{i1}$] and [$t = T$, $y_{iT}$] and the line connecting the inner point [$t = t$, $y_{iT}$], for which it holds that $t > 1$ and $t < T$, to the first measurement [$t = 1$, $y_{i1}$]. For the *i*-th object, the calculation of the radian angle for the inner point $t$, where $t > 1$ and $t < T$, can be done in the following way:

$$\triangleleft T1t = \arccos\left(\frac{(D_{iT} - D_{i1}) \cdot (D_{it} - D_{i1}) + (y_{iT} - y_{i1}) \cdot (y_{it} - y_{i1})}{\sqrt{(D_{iT} - D_{i1})^2 + (y_{iT} - y_{i1})^2} \cdot \sqrt{(D_{it} - D_{i1})^2 + (y_{it} - y_{i1})^2}}\right), \qquad (7)$$

$t = 2, ..., T-1$. The maximum angle in radians is selected as the clustering variable from $T-2$ of calculated angles $\triangleleft T1t$. The minus sign is assigned to the selected maximum angle if the slope of the tangent line passing through peripheral measurements is larger than that running through the selected inner *t* and first measurement. Otherwise, the maximum angle remains positive, i.e. the slope of the tangent line passing through peripheral measurements is smaller than that running through *t* and the first measurement. This clustering variable is referred to as $\max \triangleleft$ hereinafter.

The appended supplementary materials provide an example of the calculation of clustering variables.

### 2.2 Static clustering algorithm

The features representing the objects (trajectories) extracted in the first step (Section 2.1) are clustered in the second step described in Section 2.2. Based on the simulation study (Sobíšek, 2017), the Ward's hierarchical clustering method (Ward, 1963) applied to the Euclidean distance matrix (of seven extracted features) was chosen as the most appropriate, the effectiveness of this appproach to micro-panel data clustering being confirmed by Ferreira and Hitchcock (2009).

# 3.     Simulation study description

We compare the clustering performance of the proposed algorithm with two alternative algorithms using a simulation study. In the present study, four different types of artificial data sets have been generated, simulating real medical research micro-panel data.

## 3.1 Technical specification

In the simulation study comparing particular clustering algorithms, 10,000 repetitions were performed for four data sets – balanced and unbalanced, both low- and high-noise ones, respectively. The differences between the data sets consisted of different input parameter settings such as estimated regression parameters, variability or file size. Three different clustering algorithms were applied to these artificial data.

The overall simulation has three phases. In the first step, sample panel data were generated, based on the selected parameters (see Section 3.2). In the second step, individual clustering algorithms were gradually applied to these data. The first algorithm employed is the one from the mixAK package, belonging to the model-based category. Having determined the optimal number of mixed components, i.e. the number of clusters, the above algorithm included panel data into these clusters. This "optimal" number of clusters was then used as an input parameter to another algorithm based on the *K*-means clustering, modified for longitudinal data and included in the KML package. Finally, there is our CluMP algorithm. During the simulation, the estimated parameters were stored in clusters for a single algorithm. In the third step, the clustering results of these algorithms were compared applying the evaluation criteria (see Section 3.3).

The simulation was performed using the statistical software R; the simulation code is available upon request to the corresponding author.

## 3.2 Data description

To compare clustering methods, we generated some data sets based on multiple sclerosis research, using Uher et al. (2017) data on pathological cut-offs of global and regional brain volume loss in several multiple sclerosis phenotype groups and healthy controls. To get the biomarker data, mixed effects models are applied to magnetic resonance imaging (MRI), allowing for the estimation of the linear trend and variability of the atrophy of selected brain structures. The estimated (fixed and random) parameters of linear models for the corpus callosum relative change are employed to obtain corresponding artificial data. We have generated four data sets representing real specific MRI study designs (i.e. balanced and unbalanced designs with low and high noise, respectively) to assess the clustering performance under the four different conditions. Within each set, we consider two groups of data – patients with multiple sclerosis (MS) and healthy controls (HCs) –, yielding individual trajectories for each scenario. Tables 3.2.1 and 3.2.2 present regression parameters that were used to generate high-noise and low-noise data.

|  | $b_0$ | $b_1$ | Var $u_0$ | Var $u_1$ | Corr | Var E |
|---|---|---|---|---|---|---|
| Cluster 1 (MS) | -0.0600 | -0.7400 | 0.9999 | 0.1000 | 0.0390 | 2.1015 |
| Cluster 2 (HC) | -0.3361 | -0.2000 | 0.0703 | 0.0586 | -0.0040 | 1.3677 |

Tab 3.2.1. The regression parameters for generating high-noise data

|  | $b_0$ | $b_1$ | Var $u_0$ | Var $u_1$ | corr | Var E |
|---|---|---|---|---|---|---|
| Cluster 1 | -0.0600 | -0.7400 | 0.9999 | 0.0100 | 0.0120 | 0.1000 |
| Cluster 2 | -0.3361 | -0.2000 | 0.0703 | 0.0100 | -0.0020 | 0.1000 |

Tab 3.2.2. The regression parameters for generating low-noise data

where $b_0$ and $b_1$ denote the intercept and slope of the fixed effect, Var $u_0$ and Var $u_1$ stand for variance of the intercept and slope of the random effect, Corr is the correlation between the intercept and slope of the random effect and Var E indicates residual variance.

In the case of unbalanced study, the data were generated for 800 subjects, the two clusters being represented by 75 and 25 percent, respectively. As regards the balanced study, the data were yielded for 200 subjects, individual clusters being equally represented. The code used to generate artificial data is available in the appendix and all four data files are shown in Figure 3.2.1.

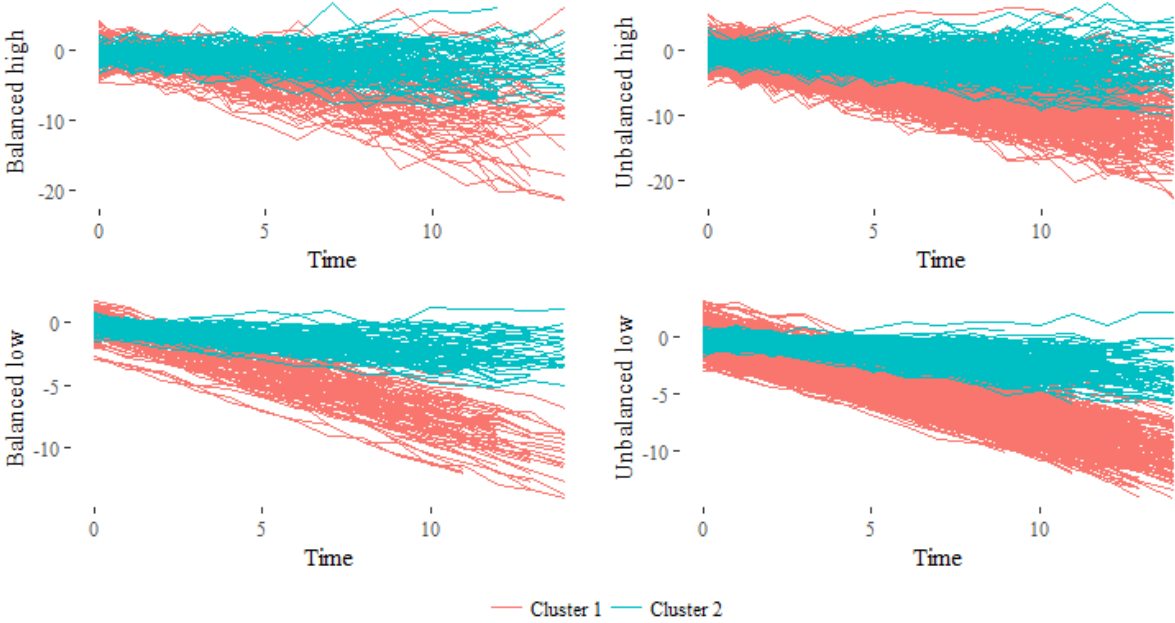

Figure 3.2.1: Artificial data sets profiles

### 3.3 Evaluation criteria

The clustering capability of algorithms has been evaluated applying four different, often used criteria. Dunn and Silhouette indices (Dunn, 1974; Rousseeuw, 1987, respectively) assess the performance of clustering based on the properties (within-cluster homogeneity and between-cluster heterogeneity) of the estimated clusters, while the Rand index (Rand, 1971) and its adjusted version (Rand, 1971; Hubert and Arabie, 1985, respectively) measure the agreements between the estimated and real (known) data groups. A more detailed description and sample calculation of the adjusted Rand index is provided by Yeung and Ruzzo, (2001). To compare clustering algorithms, the Rand index and its adjusted version were employed, e.g. in Wagner and Wagner (2007), while Dunn a Silhouette indices were used for the same comparative purpose by, for instance, Saitta et al. (2007).

## 4. Unbalanced data results

### 4.1 Results for unbalanced high-noise data

Figure 4.1.1 illustrates the distribution of values of individual indices calculated on the basis of all three algorithm clusterings for unbalanced high-noise data, also showing their median lines.

For all the indices, it holds that the higher the value, the better the clustering results. According to the Rand index and its adjusted variant, the CluMP algorithm achieves the highest and mixAK the lowest ranking, KML being placed in the middle. Silhouette and Dunn indices indicate a slightly different order of the most to the least appropriate methods – namely KML, CluMP and mixAK. Values of medians, means, standard deviations and the lower (1st) and upper (3rd) quartiles for particular indices are listed in Tables 4.1.1 and 4.1.2.

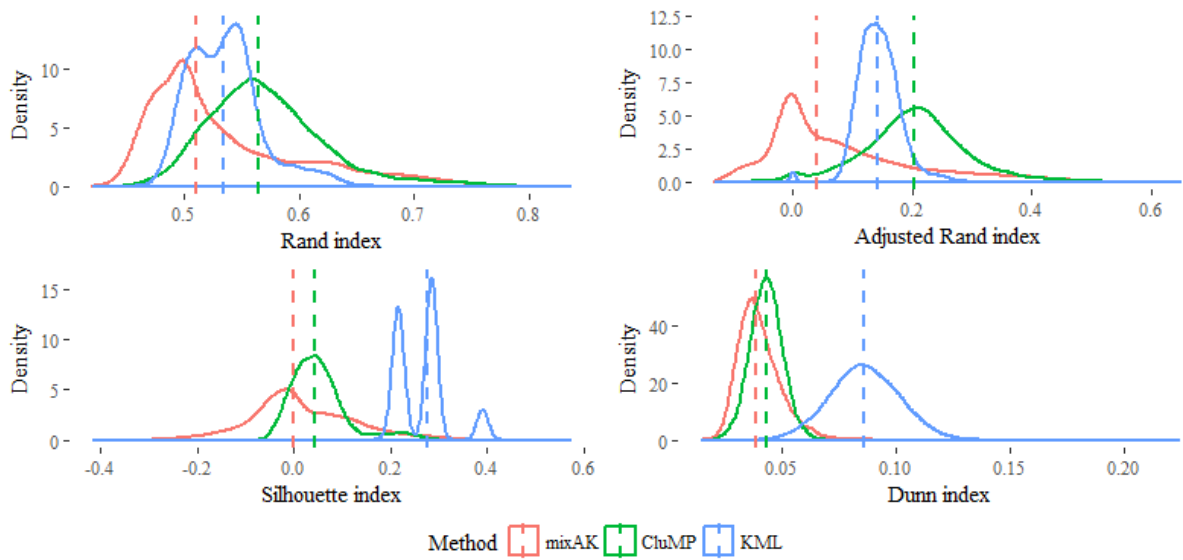

Fig. 4.1.1: Histograms for unbalanced high-noise data

|  | Rand index | | | Adjusted Rand index | | |
|---|---|---|---|---|---|---|
|  | mixAK | CluMP | KML | mixAK | CluMP | KML |
| No. of vals. | 10431 | 10431 | 10431 | 10431 | 10431 | 10431 |
| Median | 0.5101 | **0.5650** | 0.5336 | 0.0415 | **0.2050** | 0.1419 |
| Mean         | 0.5333 | **0.5712** | 0.5348 | 0.0772  | **0.2056** | 0.1450 |
| Std. dev.    | 0.0665 | 0.0517  | 0.0306 | 0.1212  | 0.0867  | 0.0335 |
| 1st quartile | 0.4866 | 0.5365  | 0.5115 | -0.0059 | 0.1551  | 0.1212 |
| 3rd quartile | 0.5643 | 0.5975  | 0.5517 | 0.1344  | 0.2534  | 0.1647 |

Tab 4.1.1. Descriptive statistic for unbalanced high-noise data: Rand and adjusted Rand indices

|              | **Silhouette index** | | | **Dunn index** | | |
|--------------|-------|-------|-------|-------|-------|-------|
|              | mixAK | CluMP | KML   | mixAK | CluMP | KML   |
| No. of vals. | 10431 | 10431 | 10431 | 10431 | 10431 | 10431 |
| Median       | -0.0002 | 0.0424 | **0.2757** | 0.0389 | 0.0432 | **0.0861** |
| Mean         | 0.0162  | 0.0510 | **0.2657** | 0.0404 | 0.0433 | **0.0863** |
| Std. dev.    | 0.1084  | 0.0600 | 0.0528 | 0.0099 | 0.0071 | 0.0151 |
| 1st quartile | -0.0505 | 0.0115 | 0.2181 | 0.0338 | 0.0385 | 0.0761 |
| 3rd quartile | 0.0830  | 0.0755 | 0.2894 | 0.0455 | 0.0479 | 0.0966 |

Tab 4.1.2. Descriptive statistic for unbalanced high-noise data: Silhouette and Dunn indices

Tables 4.1.1 and 4.1.2 display the highest average values (both the median and the mean) of all indices for each algorithm.

### 4.2 Results for unbalanced low-noise data

Figure 4.2.1 shows distributions of particular index values for unbalanced data with low noise along with their median lines. Also, for these data sets, the Rand index and its adjusted version identified CluMP as the most effective method. The difference in median values between CluMP and the second most appropriate mixAK algorithm, however, is very subtle. Just like in the case of high-noise unbalanced data, both Silhouette and Dunn indices indicated a different order of algorithms, namely KML, CluMP and mixAK. Descriptive statistic values for individual indices are available in Tables 4.2.1 and 4.2.2.

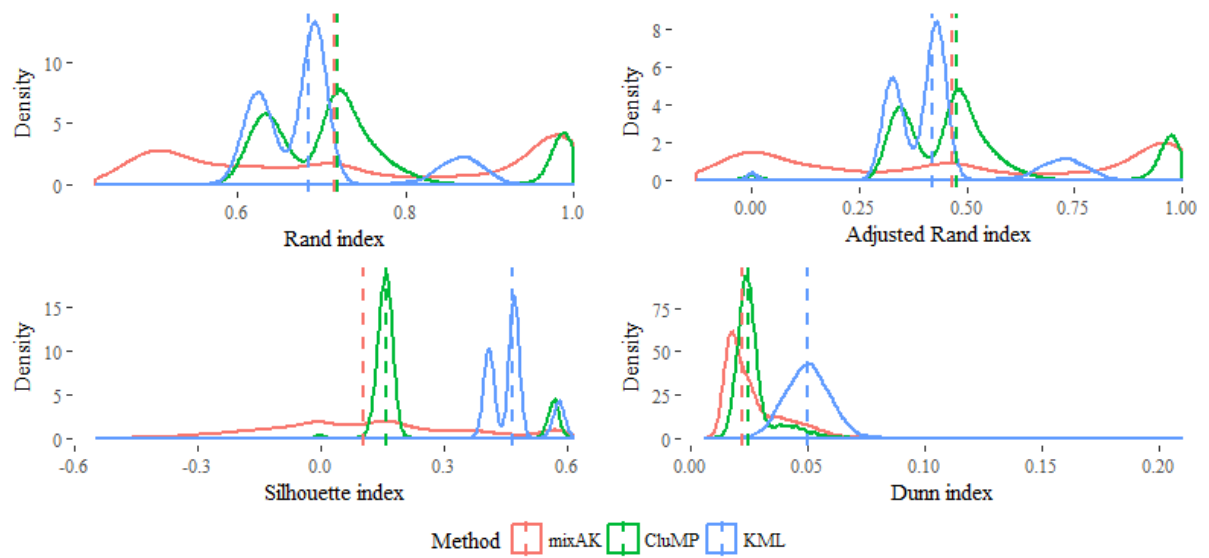

Fig. 4.2.1: Histograms for unbalanced low-noise data

|  | **Rand index** | | | **Adjusted Rand index** | | |
|---|---|---|---|---|---|---|
|  | mixAK | CluMP | KML | mixAK | CluMP | KML |
| No. of vals. | 9934 | 9934 | 9934 | 9934 | 9934 | 9934 |
| Median | 0.7185 | **0.7194** | 0.6857 | 0.4698 | **0.4776** | 0.4188 |
| Mean | **0.7454** | 0.7418 | 0.6954 | 0.4896 | **0.5257** | 0.4392 |
| Std. dev. | 0.1956 | 0.1151 | 0.0795 | 0.3959 | 0.2044 | 0.1303 |
| 1st quartile | 0.5469 | 0.6514 | 0.6358 | 0.0740 | 0.3697 | 0.3426 |
| 3rd Quartile | 0.9648 | 0.7618 | 0.7033 | 0.9263 | 0.5475 | 0.4468 |

Tab. 4.2.1. Descriptive statistic for unbalanced low-noise data: Rand and adjusted Rand indices

|  | **Silhouette index** | | | **Dunn index** | | |
|---|---|---|---|---|---|---|
|  | mixAK | CluMP | KML | mixAK | CluMP | KML |
| No. of vals. | 9934 | 9934 | 9934 | 9934 | 9934 | 9934 |
| Median | 0.1054 | 0.1625 | **0.4677** | 0.0224 | 0.0248 | **0.0500** |
| Mean | 0.1147 | 0.2192 | **0.4683** | 0.0264 | 0.0266 | **0.0500** |
| Std. dev. | 0.2342 | 0.1494 | 0.0565 | 0.0121 | 0.0079 | 0.0093 |
| 1st quartile | -0.0460 | 0.1484 | 0.4182 | 0.0178 | 0.0221 | 0.0436 |

| 3rd quartile | 0.2595 | 0.1784 | 0.4817 | 0.0318 | 0.0281 | 0.0563 |

Tab. 4.2.2. Descriptive statistic for unbalanced low-noise data: Silhouette and Dunn indices

Also, Tables 4.2.1 and 4.2.2 present the highest average values (both the median and the mean) of all indices for the respective algorithms.

## 5. Balanced data results

### 5.1 Results for balanced high-noise data

As in the previous subchapter, particular index value distributions for balanced data with high noise, along with their median lines, are displayed first (see Figure 5.1.1). Again, taking into account median values, the CluMP algorithm was determined by the Rand and adjusted Rand indices as the best, followed by KML and mixAK methods. Similar to previous data sets, both the Silhouette and Dunn indices indicated a different order of algorithms, namely KML, CluMP and mixAK. Descriptive statistic values for particular indices are given in Tables 5.1.1 and 5.1.2, respectively.

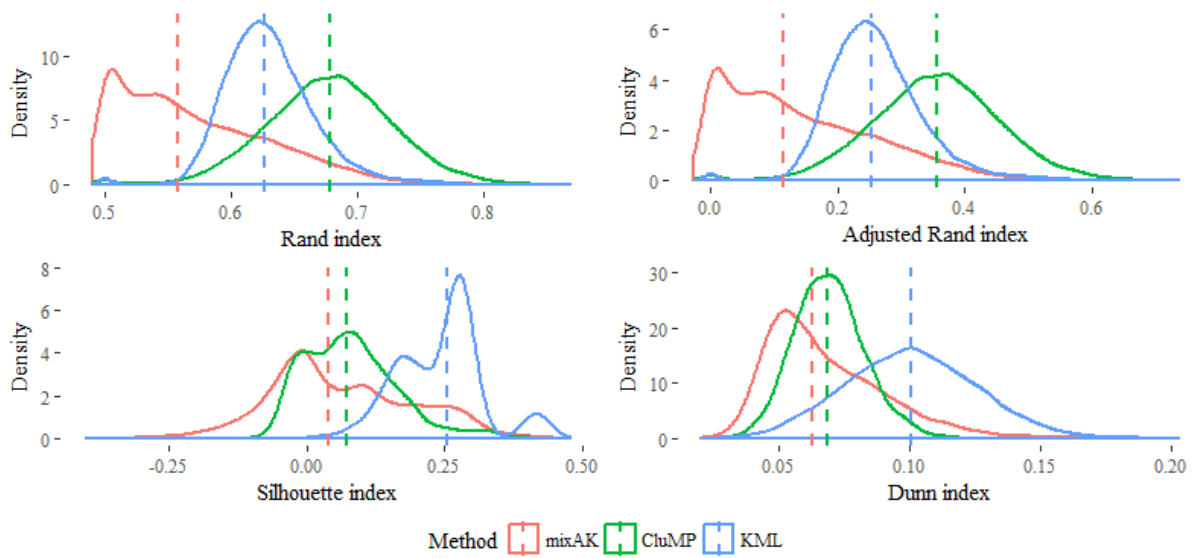

Fig. 5.1.1: Histograms for balanced high-noise data

|  | **Rand index** | | | **Adjusted Rand index** | | |
|---|---|---|---|---|---|---|
|  | mixAK | CluMP | KML | mixAK | CluMP | KML |
| No. of vals. | 9873 | 9873 | 9873 | 9873 | 9873 | 9873 |
| Median | 0.5578 | **0.6785** | 0.6266 | 0.1153 | **0.3570** | 0.2531 |
| Mean | 0.5706 | **0.6784** | 0.6303 | 0.1411 | **0.3568** | 0.2606 |
| Std. dev. | 0.0587 | 0.0479 | 0.0341 | 0.1173 | 0.0956 | 0.0680 |
| 1st quartile | 0.1028 | 0.0705 | 0.0540 | 0.8314 | 0.2680 | 0.2610 |
| 3rd quartile | 0.5213 | 0.6464 | 0.6064 | 0.0429 | 0.2925 | 0.2129 |

Tab. 5.1.1. Descriptive statistic for balanced high-noise data: Rand and adjusted Rand indices

| | Silhouette index | | | Dunn index | | |
|---|---|---|---|---|---|---|
| | mixAK | CluMP | KML | mixAK | CluMP | KML |
| No. of vals. | 9873 | 9873 | 9873 | 9873 | 9873 | 9873 |
| Median | 0.0403 | 0.0716 | **0.2533** | 0.0623 | 0.0686 | **0.1001** |
| Mean | 0.0628 | 0.0784 | **0.2422** | 0.0676 | 0.0691 | **0.1005** |
| Std. dev. | 0.1286 | 0.0832 | 0.0781 | 0.0226 | 0.0134 | 0.0250 |
| 1st quartile | 2,0471 | 1,0610 | 0.3225 | 0.3336 | 0.1934 | 0.2485 |
| 3rd quartile | -0.0281 | 0.0148 | 0.1839 | 0.0509 | 0.0599 | 0.0832 |

Tab. 5.1.2. Descriptive statistic for balanced high-noise data: Silhouette and Dunn indices

The highest average values (both the median and the mean) of all indices for each algorithm have been also presented (see Tables 5.1.1 and 5.1.2).

**5.2 Results for balanced low-noise data**

For balanced data with low noise, the results are as follows. In terms of the Rand and adjusted Rand indices, the difference between the first and second best clustering algorithms, namely CluMP and mixAK, is subtle. Silhouette and Dunn indices again indicate KML as the most efficient method, followed by mixAK and CluMP as the worst in this case. Index value distributions, including median lines, are plotted in Figure 5.2.1. Descriptive statistic values for individual indices, along with the highest averages (means and medians) of all indices for particular algorithms, are displayed in Tables 5.2.1 and 5.2.2.

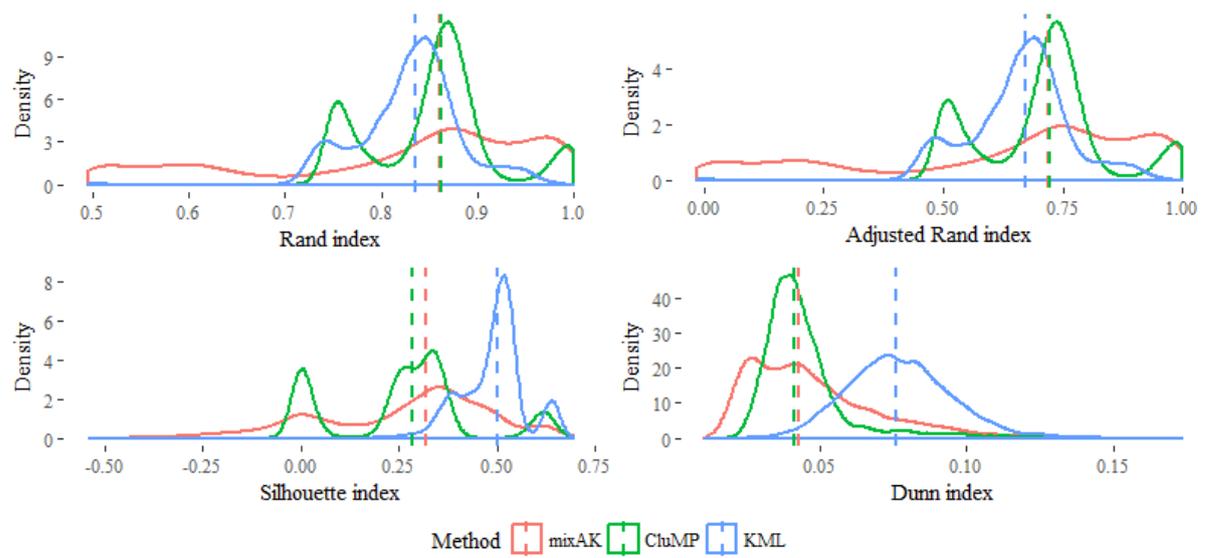

Fig.:5.2.1 Histograms for balanced low-noise data

|  | **Rand index** | | | **Adjusted Rand index** | | |
|---|---|---|---|---|---|---|
|  | mixAK | CluMP | KML | mixAK | CluMP | KML |
| No. of vals. | 9977 | 9977 | 9977 | 9977 | 9977 | 9977 |
| Median | 0.8606 | **0.8617** | 0.8353 | 0.7208 | **0.7234** | 0.6703 |
| Mean | 0.8132 | **0.8532** | 0.8304 | 0.6263 | **0.7064** | 0.6608 |
| Std. dev. | 0.1502 | 0.0663 | 0.0513 | 0.3003 | 0.1325 | 0.1026 |
| 1st quartile | 0.7117 | 0.8066 | 0.8013 | 0.4232 | 0.6138 | 0.6025 |
| 3rd quartile | 0.9305 | 0.8834 | 0.8597 | 0.8609 | 0.7666 | 0.7194 |

Tab. 5.2.1 Descriptive statistic for balanced low-noise data: Rand and adjusted Rand indices

|  | **Silhouette index** | | | **Dunn index** | | |
|---|---|---|---|---|---|---|
|  | mixAK | CluMP | KML | mixAK | CluMP | KML |
| No. of vals. | 9977 | 9977 | 9977 | 9977 | 9977 | 9977 |
| Median | 0.3167 | 0.2847 | **0.5022** | 0.0429 | 0.0413 | **0.0760** |
| Mean | 0.2651 | 0.2610 | **0.4907** | 0.0473 | 0.0445 | **0.0770** |
| Std. dev. | 0.2169 | 0.1795 | 0.0808 | 0.0219 | 0.0150 | 0.0176 |
| 1st quartile | 0.0976 | 0.1601 | 0.4436 | 0.0303 | 0.0360 | 0.0651 |
| 3rd quartile | 0.4146 | 0.3455 | 0.5322 | 0.0583 | 0.0481 | 0.0880 |

Tab. 5.2.2 Descriptive statistic for balanced low-noise data: Silhouette and Dunn indices

All types of data were assessed according to the number of clusterings equalling that of mixed components (clusters) selected by the mixAK algorithm applied as the first of the above mentioned methods. MixAK identified between one and four clusters for simulated data, with the most frequent distribution into three clusters for all types of data. The median value of the number of clusters for all types of data was also 3, the actual number of generating functions, however, being two.

## 6. Discussion

To verify the effectiveness of the three clustering methods, the simulation was used. A numerical comparison is unrealistic due to the complexity of the model approach encompassing complicated statistical properties of parameter estimates. Having conducted a simulation study, we found out that the results of our CluMP algorithm were comparable to those of the KML method. For some types of data sets, according to the selected criteria, CluMP has produced better outcomes than the other algorithms, mixAK yielding the worst results in all cases.

Compared to the proposed CluMP and KML algorithms, the model method (mixAK) has an undeniable advantage that allows for the direct estimation of trajectory parameters, representing estimated clusters and fully taking into account the uncertainty in parameter estimates. This relative disadvantage of CluMP can be overcome by estimating the regression function separately for each cluster, thus obtaining the necessary parameters.

The model approach is computationally too complex for applications to large data sets that contain tens (or hundreds) of thousands of trajectories since it does not converge to an optimal solution. This type of data can be clustered using CluMP as it is a computationally fast alternative. In order to verify this hypothesis, we simulated the length of the calculation of the three algorithms for different data sets. Time efficiency of each algorithm was measured on data sets containing 100, 200, 500, 1000, 5000 and 10000 values, calculation time depending on the size of each set. The graphical representation of time efficiency is plotted in Figure Appx.2 in the supplementary appendix.

The process of clustering based on the features of trajectories (CluMP) is not theoretically grounded in any mathematical-statistical reasoning, its principle being easy to understand for a wide range of users. Moreover, CluMP allows the user to explicitly set the number of clusters in accordance with the assumptions specified in the task. Mix AK, on the other hand, determines the number of mixed classes according to the appropriate, most comprehensive model, enabling the user to set only the maximum number of clusters created.

In the present research, the CluMP algorithm has been tested only within the univariate cluster analysis of micro-panel data, the results of simulation comparison of different approaches providing relevant information only for the above mentioned type of data and tasks. In a further study, we will test the wider applicability and performance of CluMP for multivariate analysis and other types of data, time series in particular.

# Appendix (supplementary materials)

# Appendix 1 Example calculation of characteristics (clustering variables)

The following example illustrates the transformation of the original Y values to the clustering variables. The original values and clustering variables are entered in a long and wide format, respectively. There are three objects (following three trajectories) in the example, each of them being measured annually for five years. The measured values are shown in Figure Appx. 1.

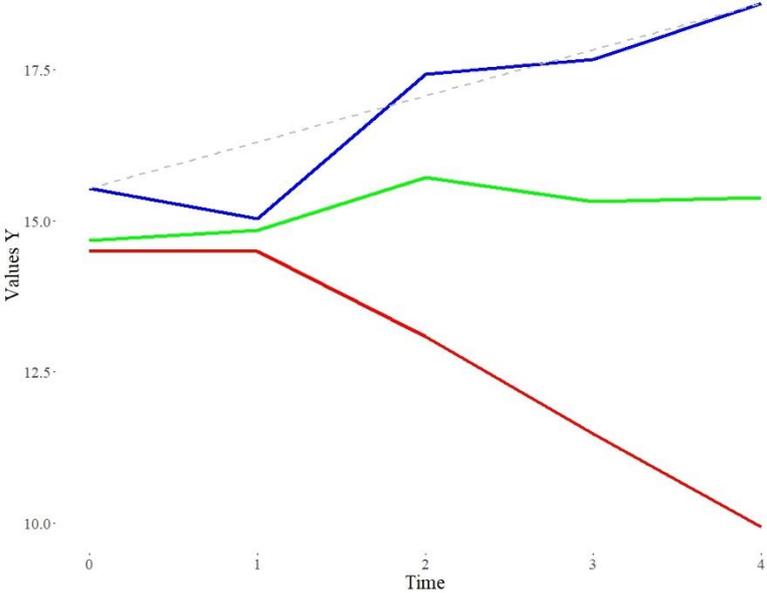

**Fig. Appx.1** Three object trajectories (plotted by blue, pink and green curves, respectively)

Apart form the measured values $y_{it}$, Table Appx. 1 provides the values of auxiliary calculations determining those of the clustering variables. The auxiliary values include $diff_{it}$, $|diff_{it}|$ and $k_{it}$, a binary variable positive change, which is either 1 for $k_{it} \geq 1$ or 0 for $k_{it} < 1$. "NA" value indicates an empty box, column $i$ identifying an object.

| $i$ | $D_{it}$ | $t$ | $y_{it}$ | $y_{i(t-1)}$ | $diff_{it}$ | $|diff_{it}|$ | $k_{it}$ | Positive change |
|---|---|---|---|---|---|---|---|---|
| 1 | 0.00 | 1 | 15.54 | NA | NA | NA | NA | NA |
| 1 | 1.00 | 2 | 15.03 | 15.54 | -0.26 | 0.26 | 0.97 | 0 |
| 1 | 2.00 | 3 | 17.42 | 15.03 | 1.20 | 1.20 | 1.16 | 1 |
| 1 | 3.00 | 4 | 17.67 | 17.42 | 0.13 | 0.13 | 1.01 | 1 |
| 1 | 4.00 | 5 | 18.59 | 17.67 | 0.46 | 0.46 | 1.05 | 1 |
| 2 | 0.00 | 1 | 14.67 | 18.59 | NA | NA | NA | NA |
| 2 | 1.00 | 2 | 14.84 | 14.67 | 0.08 | 0.08 | 1.01 | 1 |
| 2 | 2.00 | 3 | 15.71 | 14.84 | 0.44 | 0.44 | 1.06 | 1 |
| 2 | 3.00 | 4 | 15.32 | 15.71 | -0.20 | 0.20 | 0.98 | 0 |
| 2 | 4.00 | 5 | 15.38 | 15.32 | 0.03 | 0.03 | 1.00 | 1 |
| 3 | 0.00 | 1 | 14.49 | 15.38 | NA | NA | NA | NA |
| 3 | 1.00 | 2 | 14.49 | 14.49 | 0.00 | 0.00 | 1.00 | 1 |
| 3 | 2.00 | 3 | 13.09 | 14.49 | -0.70 | 0.70 | 0.90 | 0 |
| 3 | 3.00 | 4 | 11.48 | 13.09 | -0.80 | 0.80 | 0.88 | 0 |
| 3 | 4.00 | 5 | 9.94 | 11.48 | -0.77 | 0.77 | 0.87 | 0 |

**Tab. Appx. 1** Calculated values of three sample trajectories

In Table Appx. 2, the values of the clustering variables are calculated according to the formulas (1) – (7). For example, the following calculation corresponds to the first object ($i = 1$).

$$\overline{diff}_i = \frac{(2.26 + 1.20 + 0.13 + 0.46)}{4} = 0.38 ,$$

$$\text{sd}(diff_i) = \frac{(-0.26 - 0.38)^2 + (1.20 - 0.38)^2 + (0.13 - 0.38)^2 + (0.46 - 0.38)^2}{3} = 0.62 ,$$

$$\overline{|diff_i|} = \frac{(0.26 + 1.20 + 0.13 + 0.46)}{4} = 0.51 ,$$

$$\text{sd}(|diff_i|) = \frac{(0.26 - 0.38)^2 + (1.20 - 0.38)^2 + (0.13 - 0.38)^2 + (0.46 - 0.38)^2}{3} = 0.50 ,$$

$$\overline{k}_i = \sqrt[4]{\frac{18.59}{15.54}} - 1 = 0.05 \; ,$$

%pos=3/1=3.00,

$$\max \sphericalangle = \arccos\left( \frac{(D_{i5} - D_{i1}) \cdot (D_{i2} - D_{i1}) + (y_{i5} - y_{i1}) \cdot (y_{i2} - y_{i1})}{\sqrt{(D_{i5} - D_{i1})^2 + (y_{i5} - y_{i1})^2} \cdot \sqrt{(D_{i2} - D_{i1})^2 + (y_{i2} - y_{i1})^2}} \right) =$$

$$= \arccos\left( \frac{(4-0) \cdot (1-0) + (18.59 - 15.54) \cdot (15.03 - 15.54)}{\sqrt{4^2 + (18.59 - 15.54)^2} \cdot \sqrt{1^2 + (15.03 - 15.54)^2}} \right) = 1.13.$$

The angle calculated for $t = 2$, is the one between the lines connecting 1st with 5th and 1st with 2nd measurements in time. The calculation confirms, what is also evident from Figure 7.1, that just for $t = 2$, the angle with the line connecting 1st and 5th measurement (plotted in black) is the largest. It is also obvious from the Figure that the line connecting 1st with 2nd point to the right of $t = 1$ is located below the one connecting 1st with 5th measurement. The slope of the former line is smaller (–0.51) than that of the latter (0.76). For this reason, the value $\max \sphericalangle$ is negative, equalling –1.13.

| $i$ | $\overline{diff}_i$ | sd($diff_i$) | $\overline{|diff_i|}$ | sd($|diff_i|$) | $\overline{k}_i$ | %pos | max $\sphericalangle$ |
|---|---|---|---|---|---|---|---|
| 1 | 0.38 | 0.62 | 0.51 | 0.48 | 0.05 | 3.00 | –1.13 |
| 2 | 0.09 | 0.26 | 0.19 | 0.18 | 0.01 | 3.00 | 0.30 |
| 3 | –0.57 | 0.38 | 0.57 | 0.38 | –0.09 | 0.33 | 0.85 |

**Tab. Appx. 2** Suggested clustering variables representing the dynamics of three sample trajectories

Time efficiency of all three applied alghorithms.

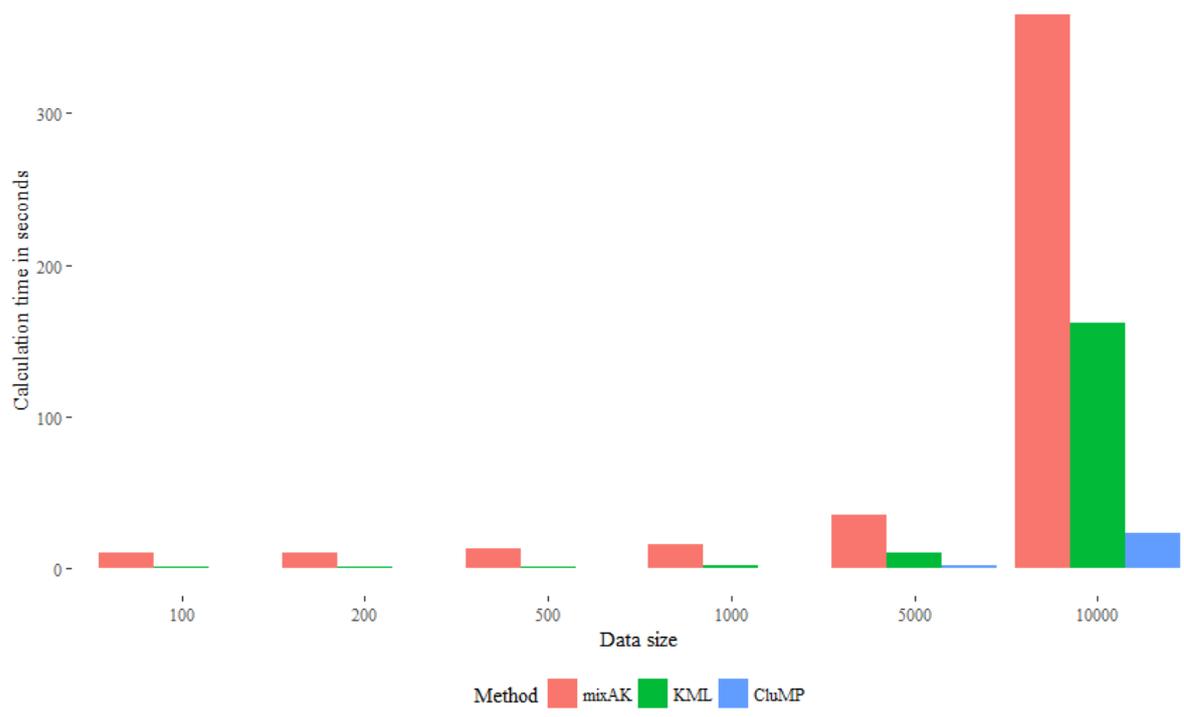

**Fig. Appx. 2** Time efficiency of each alghorithm